\documentclass[lettersize,journal]{IEEEtran}
\usepackage{amsmath,amsfonts}
\usepackage{algorithmic}
\usepackage{array}
\usepackage[caption=false,font=normalsize,labelfont=sf,textfont=sf]{subfig}
\usepackage{amsmath}
\usepackage{amssymb}
\usepackage{amsfonts}
\usepackage{booktabs, makecell, multirow, tabularx}
\usepackage{textcomp}
\usepackage{stfloats}
\usepackage{url}
\usepackage{verbatim}
\usepackage{graphicx}
\usepackage{xcolor}
\usepackage{spacingtricks}
\usepackage{bbm}
\usepackage{multirow}
\usepackage{makecell}
\usepackage{graphicx}
\usepackage{amsmath}
\usepackage{booktabs}
\usepackage{multirow}
\usepackage{array}
\usepackage{amsfonts}
\usepackage{bbm} 
\usepackage{xcolor}
\usepackage{mdframed}
\usepackage{makecell}
\usepackage[normalem]{ulem}
\usepackage{graphicx}
\usepackage{subcaption} 

\usepackage{amsmath,amsfonts,bm}

\usepackage{mdframed}

\usepackage{lipsum}

\newcommand{\nk}{\kern-0.1em}

\def\BibTeX{{\rm B\kern-.05em{\sc i\kern-.025em b}\kern-.08em
    T\kern-.1667em\lower.7ex\hbox{E}\kern-.125emX}}
\usepackage{balance}
\begin{document}
\title{
Fully Test-Time rPPG Estimation via Synthetic Signal-Guided Feature Learning
} 


\author{Pei-Kai Huang, Tzu-Hsien Chen, Ya-Ting Chan, Kuan-Wen Chen, Chiou-Ting Hsu, ~\IEEEmembership{Senior Member,~IEEE} 
\\
Department of Computer Science, National Tsing Hua University, Hsinchu, Taiwan
}


\maketitle

\begin{abstract} 
Many remote photoplethysmography (rPPG) estimation models have achieved promising performance in the training domain but often fail to accurately estimate physiological signals or heart rates (HR) in the target domains.  
Domain generalization (DG) or domain adaptation (DA) techniques are therefore adopted during the offline training stage to adapt the model to either unobserved or observed target domains by utilizing all available source domain data.
However, in rPPG estimation problems, the adapted model usually encounters challenges in estimating target data with significant domain variation, such as different video capturing settings, individuals with different HR ranges, or unbalanced HR distributions. 
In contrast, Test-Time Adaptation (TTA) enables the model to adaptively estimate rPPG signals in various unseen domains by online adapting to unlabeled target data without referring to any source data.
In this paper, we first establish a new TTA-rPPG benchmark that encompasses various domain information and HR distributions to simulate the challenges encountered in real-world rPPG estimation. 
Next, we propose a novel synthetic signal-guided rPPG estimation framework to address the forgetting issue during the TTA stage and to enhance the adaptation capability of the pre-trained rPPG model.
To this end, we develop a synthetic signal-guided feature learning method by synthesizing pseudo rPPG signals as pseudo ground truths to guide a conditional generator in generating latent rPPG features.
In addition, we design an effective spectral-based entropy minimization technique to encourage the rPPG model to learn new target domain information. 
Both the generated rPPG features and synthesized rPPG signals prevent the rPPG model from overfitting to target data and forgetting previously acquired knowledge, while also broadly covering various heart rate (HR) distributions.
Our extensive experiments on the TTA-rPPG benchmark show that the proposed method achieves superior performance and outperforms previous DG and DA methods across most protocols of the benchmark. 
\end{abstract}

\section{Introduction}

\begin{figure} 
    \begin{minipage}[t]{1.\columnwidth}
      \centering{ %
      \includegraphics[width=\columnwidth]{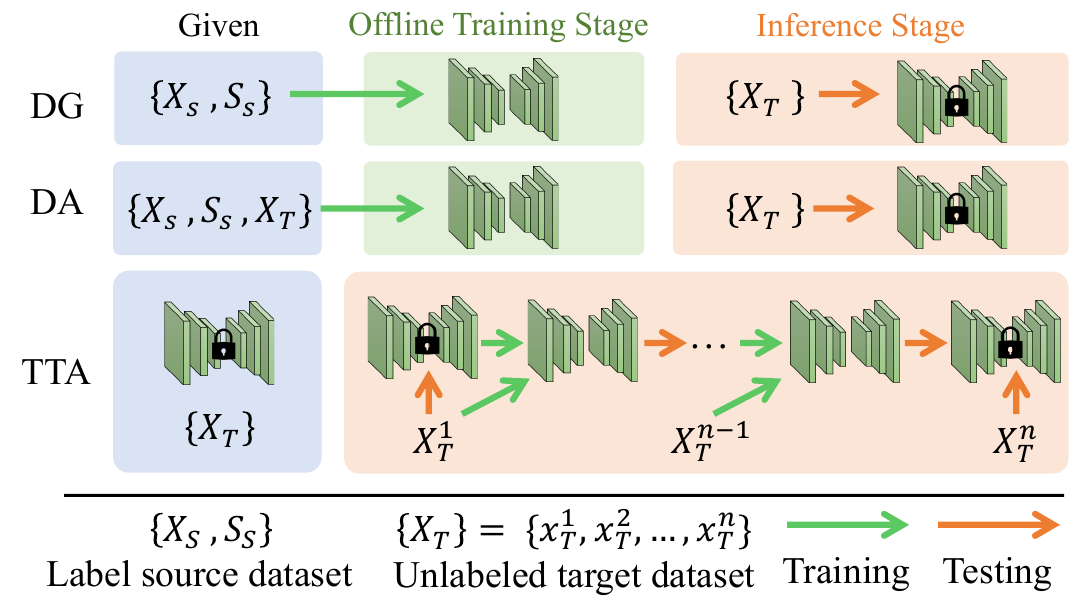} 
      } 
    \end{minipage}   
\caption{ 
 Illustration of different cross-domain scenarios in rPPG estimation: (a) Domain Generalization (DG), (b) Domain Adaptation (DA), and (c)  Test-Time Adaptation (TTA).  
Note that, unlike DG and DA, TTA includes no offline training stage but involves adapting the model in the inference stage, which consists of one  online training (adaptation) process and one online testing process.
}
 \label{fig:TTA}   
\end{figure}

Remote photoplethysmography (rPPG) is a contactless technology of estimating physiological signals and heart rate (HR) related information through
measuring subtle chrominance changes reflected on facial skin. To generalize the rPPG estimation model to various test domains, many recent methods \cite{hsieh2022augmentation, lu2023neuron, du2023dual} have adopted domain generalization (DG) or domain adaptation (DA) approaches to adapt the model trained on source domain to either unobserved or observed target domain.
As shown in Figure~\ref{fig:TTA} (a), DG technique \cite{hsieh2022augmentation, lu2023neuron} aims to learn a generalized rPPG model from multiple source domains during the training stage and then use this offline trained model in the inference stage.
In Figure~\ref{fig:TTA} (b), if the target data is also available, then DA technique \cite{du2023dual} can be readily applied to transfer the source knowledge into the target domain.
In the two settings depicted in Figure~\ref{fig:TTA}, both DG and DA techniques utilize the fixed and offline trained models in the inference stage.
Since target data often encompass diverse domain information different from the source domain, these fixed models deem impractical to resolve the unknown or unseen test domain information, such as specific video capturing settings, different individuals, varying age ranges or HR distributions.
Moreover, the pre-adapted models in the offline training stage usually yield unstable estimation due to potential inaccuracies in the ground truth labels of source training datasets \cite{hsieh2022augmentation}.

Unlike DG and DA, fully Test-Time Adaptation (TTA) addresses a more practical scenario where only an off-the-shelf model is accessible without access to source domain data. Thus, TTA focuses on adapting this off-the-shelf model directly to unlabeled target data during the inference stage, as illustrated in Figure~\ref{fig:TTA} (c).
Many recent TTA methods \cite{wang2021tent, lin2023video} have demonstrated promising performance improvements in various computer vision tasks.
For example, the authors in \cite{wang2021tent} proposed adapting the model to learn target domain information by minimizing entropy to update the affine parameters of batch normalization (BN) layers for image classification. 
In \cite{lin2023video}, the authors proposed aligning the feature distribution between the source and target domains to address common distribution shifts for video action recognition.
In addition, to effectively adapt the modal to target domain, the authors in \cite{lim2022ttn,park2023label,choi2022improving} 
including an additional post-training stage (before the inference stage within  Figure~\ref{fig:TTA}) to obtain more information. 
While TTA has shown promising potential in many tasks, it still remains unexplored in the field of rPPG estimation.
 
Compared to other computer vision tasks, rPPG estimation relies on subtle color changes for signal extraction and faces more challenges in TTA setting.
Since the target data is collected from multiple environments and devices, different facial videos usually exhibit significant distribution discrepancies between the source and target domains.
Therefore, the first challenge is fine-tuning the rPPG model to adapt to the new domain information of the target domain.
The second challenge concerns the unlabeled target data problem.
When adapting a rPPG model to unlabeled target data, TTA encounters the same challenge as DA of learning rPPG signals from the unlabeled target data to enable model adaptation.
In addition, unlike other image or video classification tasks that primarily deal with known classes, rPPG models may encounter heart rate ranges unseen in the source training domain during the adaptation process.
Therefore, when encountering target data containing new domain information and unseen heart rate distributions, TTA for rPPG estimation becomes doubly challenging.

This paper addresses the above-mentioned challenges of TTA in rPPG estimation and proposes a novel synthetic signal-guided rPPG estimation framework to effectively adapt the rPPG model to the target domain during the inference stage.
First, to address the problem regarding domain information, we propose an effective spectral-based entropy minimization by minimizing the entropy of the power spectrum densities (PSDs) derived from estimated rPPG signals.
Next, to overcome the challenge of unseen HR distributions, we propose a novel synthetic signal-guided generative feature learning approach.
In particular, we propose synthesizing rPPG signals within human HR constraint (i.e., the frequency interval between 0.66 Hz and 4.16 Hz \cite{sun2022contrast}) as pseudo ground-truths to guide a conditional generator to generate nontrivial latent rPPG features.
To store these synthesized rPPG signals and generated rPPG features, we design a hierarchical memory bank for each frequency.
During the inference stage of TTA, these generated rPPG signals and features are utilized to guide the rPPG models in learning unseen HRs. 
Furthermore, to adapt to unlabeled target data, we propose an effective adaptive feature alignment to align the feature distributions between the target rPPG features and the generated rPPG features. 
This alignment facilitates a more effective adaptation of the rPPG model to the characteristics of the target data during inference.

To evaluate the proposed framework on TTA setting, we further propose a comprehensive rPPG benchmark: TTA-rPPG, i.e., Test Time Adaptation for rPPG estimation. 
In the TTA-rPPG benchmark, we design two challenging real-world scenarios to simulate the realistic scenario of test-time adaption: 
(1) new domain information meets seen HR distribution and 
(2) new domain information meets unseen HR distribution. 
Our experimental results on the TTA-rPPG benchmark not only verify the effectiveness of the proposed framework in the TTA setting but also open up many potential directions for future research.  

Our contributions are summarized as follows:

\begin{itemize}
\item  We propose a new benchmark, TTA-rPPG, which covers various domain information and HR distributions to closely simulate the real-world scenario. To the best of our knowledge, this is the first work that addresses test-time adaptation for rPPG estimation.  

\item  
Under the heart rate constraint, we synthesize rPPG signals as pseudo ground-truths and then propose a synthetic signal-guided generative feature learning  to generate latent rPPG features for enhancing the adaptation capability of rPPG models.

\item  
Extensive experiments demonstrate that the proposed framework achieves superior performance in online adapting to both seen and unseen HR distributions across new domains. 

\end{itemize}

\section{Related Work}
\label{sec:Related Work}

\subsection{Test-time adaptation}
 
Test-time Adaptation (TTA) focuses on using the testing data to online fine-tune the off-the-shelf models to adapt to the target domain during the inference stage.
Recent TTA methods generally fall into two categories: non-fully and fully test-time adaptation. 
 
Non-fully test-time adaptation aims to adapt the off-the-shelf models to target domain by using the auxiliary source statistics during the inference stage.
For example,  the authors in  \cite{lim2022ttn,park2023label,choi2022improving} proposed accessing the source data in the post-training stage (after the pre-training stage, before the inference stage) to obtain the source statistics, and then incorporating these source statistics to facilitate the adaptation process of TTA.
Next, the authors in \cite{zhang2023domainadaptor} proposed adaptively mixing the statistics between source and target domains to increase the diversity of the statistics, and then using the mixed statistics to enhance TTA.
In addition, in \cite{lin2023video}, the authors proposed to align online estimates of test set statistics towards the training statistics for video action recognition. 
However, obtaining the source data and its statistics is not easily feasible in real-world scenarios.

\begin{figure*} [t]
    \begin{minipage}[b]{1.0\linewidth}
      \centering{ 
      \includegraphics[width=1.0\linewidth]{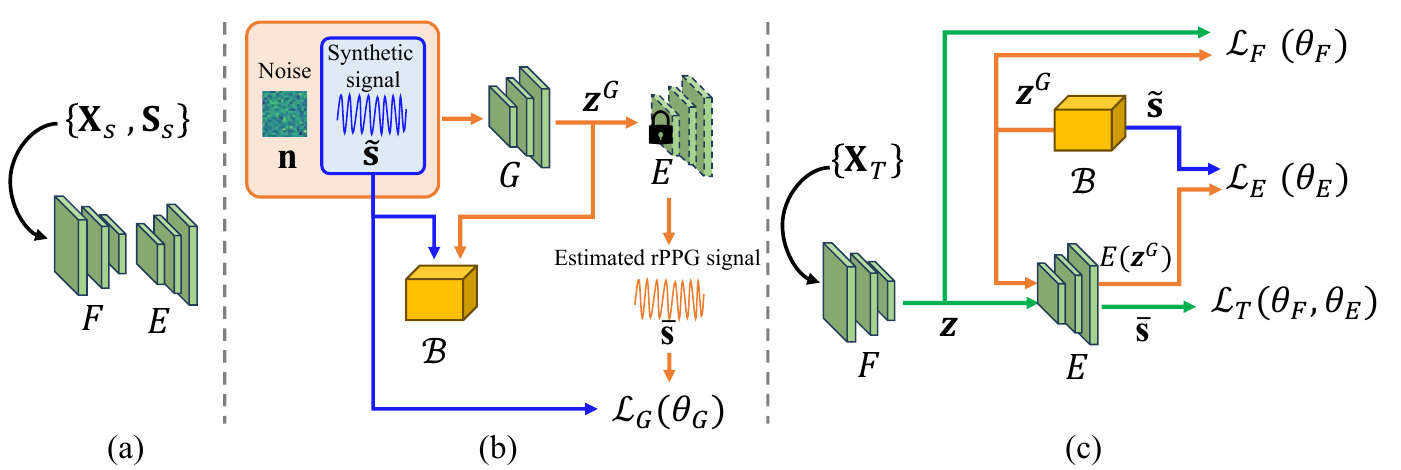}
      }
    \end{minipage} 
\vspace{-20px}
\caption{  
The proposed framework consists of one feature extractor $F$, one rPPG estimator $E$, and one latent rPPG feature generator $G$. (a) The pre-training stage: the off-the-shelf rPPG model $T=F\circ E$ is pretrained on the source training data $\{\mathbf{X}_s,\mathbf{S}_s\}$. 
(b) The post-training stage: the sampled Gaussian noise $\mathbf{n}$ and the synthetic rPPG signal $\widetilde{\mathbf{s}}$ are used to guide the generator $G$ to generate latent rPPG features $ \mathbf{z}^G$.   
(c) The inference stage of test-time adaptation: we first adapt $T$ to the incoming batch of target sample $\{ \mathbf{X}_T\}$ to alleviate the uncertainty of the estimated rPPG signals, and then use the generated paired $ \mathbf{z}^G$ and $\widetilde{\mathbf{s}}$ to adapt $T$ to target domain.
}
 \label{fig:framework}   
\end{figure*}  
 
Compared to obtaining the source data and its statistics, fully test-time adaptation is more feasible, as fully test-time adaptation requires only the acquisition of off-the-shelf models.
Therefore, the authors in \cite{wang2021tent} pioneered  fully test-time adaptation by adapting the off-the-shelf models to the target domain without the need for auxiliary source statistics.
Next, in \cite{zhao2023delta}, the authors proposed to rectify inaccurate normalization statistics to address the issue of class bias.
Furthermore, the authors in \cite{boudiaf2022parameter} proposed adapting concave-convex procedure to optimize the models for fully test-time adaptation.
Considering the potential of fully test-time adaptation, we firmly believe that exploring fully test-time adaptation would boost the development of the field in rPPG estimation.

\subsection{rPPG estimation} 
 
Remote photoplethysmography (rPPG) estimation aims to derive pulse signals from facial videos by capturing and analyzing the subtle changes in skin color for heart rate measurement. 
Since rPPG signals are subtle and fragile, rPPG estimation is a challenging task.
rPPG estimation often performs well in the source training domain but fails to generalize effectively to the target testing domain due to new domain information and unseen heart rate ranges. 
In addition, due to the potential inaccuracies of the ground truth rPPG signals within source training data \cite{hsieh2022augmentation}, the learned rPPG models may exhibit unstable estimations. 
In earlier methods \cite{niu2019rhythmnet, tsou2020multi, lee2020meta}, the authors  focused on improving intra-domain performance but rarely addressed the cross-domain issue.
These rPPG models easily overfit to the source training domain and fail to generalize to the target testing domain due to significant domain shifts.
To address the cross-domain issue, many recent rPPG estimation methods have been developed to adopt domain generalization (DG) and domain adaptation (DA) techniques. 
For example,  the authors in \cite{lu2023neuron,wang2023hierarchical} proposed adopting DG technique to address the cross-domain issue for rPPG estimation.
In particular, the authors in \cite{lu2023neuron} proposed to maximize  the coverage of the rPPG feature space to improve the generalization capacity of rPPG extraction. 
In \cite{wang2023hierarchical}, the authors
proposed to learn a hierarchical style-aware feature disentanglement to extracting domain-invariant and instance-specific rPPG features to enhance rPPG estimation.
In addition,  the authors in  \cite{du2023dual} proposed adopting domain adaptation (DA) technique to synthesize target noise to enhance the source domain training to achieve the noise reduction of rPPG estimation.

\section{Methodology}  
\label{sec:Proposed Method} 
 
In this paper, we aim to tackle the test-time adaptation (TTA) problem in rPPG estimation. To this end, we propose a novel synthetic signal-guided generative feature learning method that adapts an off-the-shelf rPPG model during the inference stage. We particularly focus on adapting to domains with unbalanced HR distribution or with HR ranges different from those in the source domain. 
 
 
Figure~\ref{fig:framework} illustrates the three stages in the proposed Test-Time rPPG estimation: (a) pre-training stage, (b) post-training stage, and (c) inference stage. Below, we provide a detailed description of these three stages. 


\subsection{Pre-training stage}   
To simulate the pre-training stage in the TTA scenario, as illustrated in Figure~\ref{fig:framework} (a), we use a simple rPPG model $T=E\circ F$ composed of a feature extractor $F$ and an rPPG estimator $E$. That is, for a facial video  $\mathbf{x}$, we obtain its estimated rPPG signals $\bar{\mathbf{s}}$ by,  

\begin{equation}
    \bar{\mathbf{s}} = T(\mathbf{x}) = E(F(\mathbf{x}))= E(\mathbf{z}).
\label{eqn:modelT}
\end{equation}

\noindent 
where $\mathbf{z} = F(\mathbf{x})$ is the latent rPPG feature of $\mathbf{x}$. Note that, any existing network architecture can be adopted for $T$. In our experiments, we use PhysNet from \cite{yu2019physnet} to build $T$ and will also compare with other network architectures in our ablation study.
We train the model $T$ using the source training data $\{\mathbf{X}_s,\mathbf{S}_s\}$  (where $\mathbf{X}_s$ represents the source facial videos and $\mathbf{S}_s$ denotes the corresponding ground truth rPPG signals) with negative Pearson correlation loss \cite{tsou2020multi, tsou2020siamese}.

\begin{figure}  
    \centering
    \begin{tabular}{c c}
        \begin{minipage}{0.85\columnwidth}
            \includegraphics[width=\linewidth]{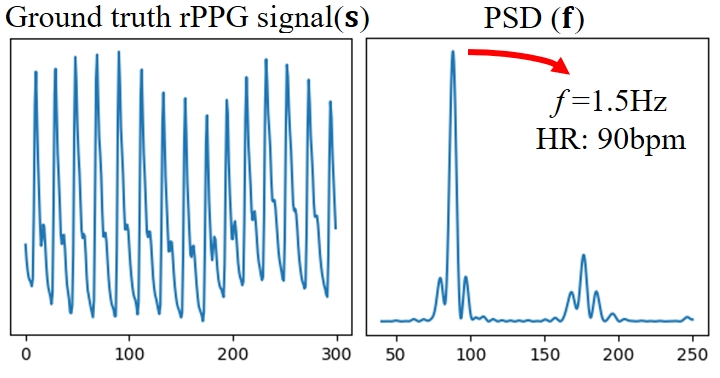}
        \end{minipage} \\ (a)
        \\ 
        \begin{minipage}{0.85\columnwidth}
            \includegraphics[width=\linewidth]{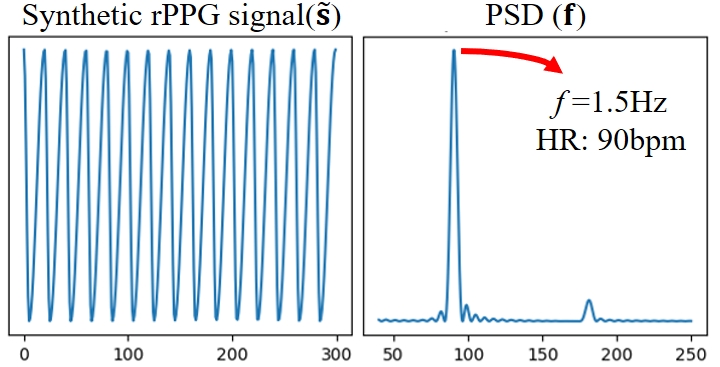}
        \end{minipage} \\ (b)
        \\ 
    \end{tabular}
\caption{  
Examples of (a) a ground truth rPPG signal and its corresponding Power Spectral Density (PSD), and (b) a synthetic rPPG signal and its corresponding PSD. 
} 
\label{fig:rPPG signals}  
\end{figure}

\subsection{Post-training stage: synthetic signal-guided feature learning
} 
In this stage, as illustrated in Figure~\ref{fig:framework} (b), we first synthesize a set of pseudo rPPG signals $\widetilde{\mathcal{S}} = \{\widetilde{\mathbf{s}}\}$  and then incorporate a generative module $G$ to produce latent rPPG features $\mathbf{z}^G$ using these synthetic rPPG signals as pseudo ground truth. 


\subsubsection{Synthetic rPPG signals}
 
To obtain crucial supplementary rPPG information for addressing the issues of forgetting and unseen HR ranges, we synthesize a set of pseudo rPPG signals  $\widetilde{\mathcal{S}} = \{\widetilde{\mathbf{s}}\}$ that covers the entire range of human HR. 
Previous studies \cite{gideon2021way,sun2022contrast} have indicated that human HR ranges from 40 to 250 beats per minute (bpm).
 In addition, several methods \cite{gideon2021way,sun2022contrast} have demonstrated that the highest peak in power spectrum densities (PSDs) effectively identifies the frequency corresponding to the HR value, as shown in Figure \ref{fig:rPPG signals} (a).


 
Therefore, we convert the human HR range (i.e., 40 to 250 bpm) into the corresponding frequency range $f$ of $[0.66, 4.16]$ Hz and then utilize the simplified formulation from \cite{niu2018synrhythm} to synthesize $\widetilde{\mathbf{s}}$ by, 
\begin{equation}
\label{eqn:syn_rPPG}
\begin{split}
\widetilde{\mathbf{s}} &= M \sin(2 \pi f t + \phi) + 0.1 \cdot M \sin(2 \pi (2f) t + \phi), 
\end{split}
\end{equation} 

\noindent  
where $\operatorname{sin}(\cdot)$ represents the sine function, $M$ denotes the magnitude randomly sampled from $[0,1]$, $t$ represents time, and $\phi$ is the phase offset randomly sampled from $[0, 2\pi]$.
Figure~\ref{fig:rPPG signals} (b) shows an example of synthesized rPPG signal corresponding to an HR of 90 bpm and $f=1.5$Hz. 
As shown in Figures~\ref{fig:rPPG signals} (a) and (b), the synthetic and ground truth rPPG signals indeed closely resemble each other. 


\subsubsection{Generative feature learning}  
After obtaining the set of pseudo rPPG signals $\widetilde{\mathcal{S}}$, we develop a synthetic signal-guided feature learning approach to generate latent rPPG features by using $\widetilde{\mathcal{S}}$ as the pseudo ground truth. 
As illustrated in Figure~\ref{fig:framework} (b), we propose a latent feature generation process, where we train a generator $G$ conditioned on a given synthetic rPPG signal $\widetilde{\mathbf{s}} \in \widetilde{\mathcal{S}}$ and a noise $\mathbf{n}$ sampled from a unit Gaussian prior, $\mathbf{n} \sim \mathcal{N}(0, I)$, to generate latent rPPG features $\mathbf{z}^G$ by, 
\begin{equation}
   \mathbf{z}^G \, \leftarrow G(\mathbf{n} | \widetilde{\mathbf{s}}; \bm{\theta}_G ).  
\label{eqn:zG}
\end{equation}
\noindent 
 
As illustrated in Figure~\ref{fig:framework} (b),  the synthetic signal-guided feature learning pipeline involves both the generator $G$ and the estimator $E$, incorporating the parameters $\bm{\theta}_G$ and $\bm{\theta}_E$, respectively.
We adopt an optimization strategy to learn $\bm{\theta}_G$ while keeping $\bm{\theta}_E$ fixed by minimizing the following loss: 
\begin{equation} \label{eqn:loss_G}
\begin{split}
\mathcal{L}_G(\bm{\theta}_G) & = \sum\nolimits_{\widetilde{\mathbf{s}} \in \widetilde{\mathcal{S}}}   1-  \rho (E(G(\mathbf{n} | \widetilde{\mathbf{s}}; \bm{\theta}_G )), \widetilde{\mathbf{s}})   \\ 
 & =  \sum\nolimits_{\widetilde{\mathbf{s}} \in \widetilde{\mathcal{S}}}   1-  \rho (E( \mathbf{z}^G), \widetilde{\mathbf{s}})  ,
\end{split}
\end{equation}
\noindent 
where $\rho(  \bar{\mathbf{s}} , \widetilde{\mathbf{s}} ) =  \frac{Cov(  \bar{\mathbf{s}} , \widetilde{\mathbf{s}} )}{\sqrt{Cov(   \bar{\mathbf{s}} , \bar{\mathbf{s}} )}\sqrt{Cov(  \widetilde{\mathbf{s}} , \widetilde{\mathbf{s}} )}} $ is the Pearson correlation between $\bar{\mathbf{s}} = E( \mathbf{z}^G)$ and $\widetilde{\mathbf{s}}$,  
and $Cov(\bar{\mathbf{s}}, \bar{\mathbf{s}})$ denotes the covariance between $\bar{\mathbf{s}}$ and $\bar{\mathbf{s}}$. 
To learn $\bm{\theta}_G$, we define the optimization problem as follows:  
\begin{equation}
   \bm{\theta}_G^* = \mathop{\mathrm{arg}\, \mathrm{min}}_{ \bm{\theta}_G} \mathcal{L}_G(\bm{\theta}_G).
\label{eqn:theta_G}
\end{equation}

\begin{figure}[t]
    \centering
    \begin{minipage}[t]{0.7\columnwidth}
      \centering{ 
      \includegraphics[width=1\columnwidth]{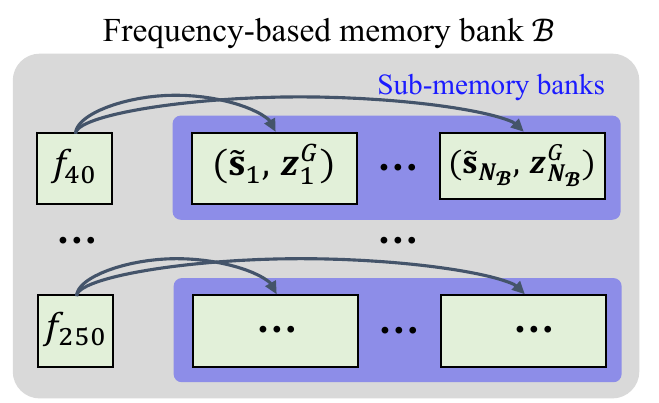}
      }
    \end{minipage}   
\caption{
Structure of the frequency-based memory bank $\mathcal{B}$.
}
 \label{fig:MB}   
\end{figure}   

\subsubsection{Frequency-based memory bank}
After establishing the paired synthetic rPPG signals and latent features $\{ (\widetilde{\mathbf{s}},\mathbf{z}^G ) \}$, we then develop a frequency-tiered memory bank $\mathcal{B}$ to store the generated these data $  (\widetilde{\mathbf{s}},\mathbf{z}^G)$, organized according to the frequencies corresponding to human HR between 40 and 250 bpm. 
As illustrated in Figure~\ref{fig:MB}, within the memory bank $\mathcal{B}$, we maintain a fixed-size sub-memory bank of size $N_{\mathcal{B}}$ for each frequency and use a First In First Out (FIFO) scheme to update the sub-memory bank.

\begin{figure}  
    \centering
    \begin{tabular}{c c}
        \begin{minipage}{0.85\columnwidth}
            \includegraphics[width=\linewidth]{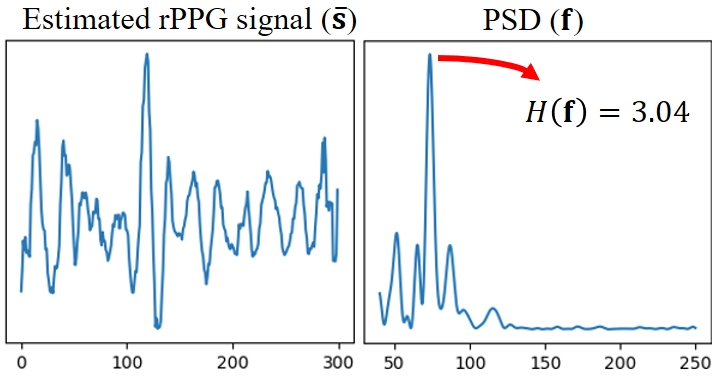}
        \end{minipage} \\ (a)
        \\ 
        \begin{minipage}{0.85\columnwidth}
            \includegraphics[width=\linewidth]{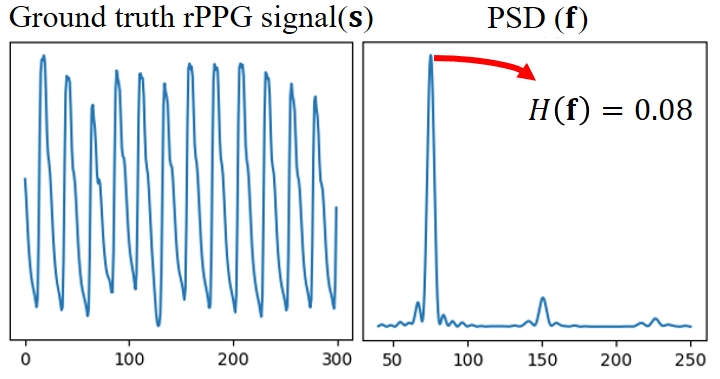}
        \end{minipage} \\ (b)
        \\ 
    \end{tabular}
\caption{  
    Examples of the entropy  $H(\mathbf{f})$ measured on the PSDs of (a) the estimated rPPG signal $\bar{\mathbf{s}}$, and (b) the ground truth rPPG signal $\mathbf{s}$. 
} 
    \label{fig:ent}
\end{figure}  
  
\subsection{Inference stage of test-time adaptation} 
Figure \ref{fig:framework} (c) illustrates the proposed online training (adaptation) process in the inference stage of TTA.
Here, we leverage the supplementary information stored in the memory bank $\mathcal{B}$ and propose two key strategies to boost adaptation capability: spectral-based entropy minimization and feature-aligned rPPG estimation.   
 
\subsubsection{Spectral-based entropy minimization} 
To minimize the potential domain discrepancy between the pre-trained rPPG model $T$ and the target data $\mathbf{X}_T$, we develop a spectral-based entropy minimization method to enhance the credibility of the estimation.
We use the examples in Figure~\ref{fig:ent} to illustrate our idea. When estimating the entropy $H(\mathbf{f})$  \cite{bercher2000estimating} on the PSDs of the estimated rPPG signal $\bar{\mathbf{s}}$ (Figure~\ref{fig:ent} (a)) and the ground truth rPPG signal $\mathbf{s}$ (Figure~\ref{fig:ent} (b)) from the same video, we observe that $\bar{\mathbf{s}}$ exhibits much higher entropy compared to $\mathbf{s}$, even though their peak PSDs are nearly identical.
To reduce this uncertainty in $\bar{\mathbf{s}}$, we incorporate a spectral-based entropy loss to improve adapting the model $T$ to the target data by,  
\begin{equation} 
\mathcal{L}_{T}(\bm{\theta}_F, \bm{\theta}_E) 
= H(\mathbf{f})
= -\sum^b_{i=a}  \mathbf{f}_i \log( \mathbf{f}_i),
\label{eq:ent}
\end{equation}

\noindent 
where $\mathbf{f}$ is the PSD of $\bar{\textbf{s}}$,
$\mathbf{f}_i$ denotes the $i$-th frequency bin within $\mathbf{f}$,  
and $a$ and $b$ represent the lower and upper bounds of the frequency (i.e., 0.66 Hz and 4.16 Hz).

\subsubsection{Feature-aligned rPPG estimation}
 
To further adapt to the characteristics of the target data $\mathbf{X}_T$, by utilizing the pairs of generated data $(\widetilde{\mathbf{s}}, \mathbf{z}^G)$ stored in the memory bank $\mathcal{B}$, we propose aligning the feature distributions between the target rPPG features and the generated rPPG features $\mathbf{z}^G$. We first define a feature extraction loss to fine-tune the feature extractor $E$ by, 
\begin{equation}
\label{eqn:loss_E}
\begin{split}
\mathcal{L}_E (\bm{\theta}_E) = \sum\nolimits_{
 \widetilde{\mathbf{s}},\mathbf{z}^G \in \mathcal{B} } 1-  \rho (E( \mathbf{z}^G; \bm{\theta}_E), \widetilde{\mathbf{s}}).
\end{split}
\end{equation}
  
\begin{figure}[t] 
    \centering 
     \includegraphics[width=0.95\columnwidth]{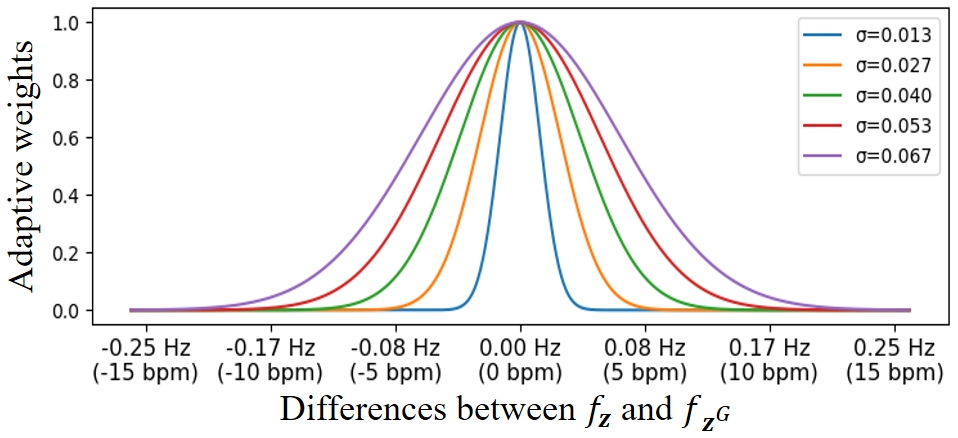} 
\caption{
Illustration of frequency-adaptive weights $e^{\frac{-1}{2}(\frac{f_ \mathbf{z} - f_{\mathbf{z}^G}}{\sigma})^2}$ for aligning rPPG features across different frequencies and HR ranges.
}
 \label{fig:sigma_range}   
\end{figure}

Next, we need to fine-tune the extractor $F$ to align the extracted rPPG features $\textbf{z} = F(\mathbf{x})$ with the generated rPPG features $\textbf{z}^G$ corresponding to similar HR values. Note that, to enhance the learning of rPPG features,  previous rPPG estimation methods \cite{gideon2021way, sun2022contrast}  proposed to dichotomize the rPPG features as either positive pairs or negative pairs based on different HR values. 
However, as noted in \cite{yang2023simper}, periodic features often show continuous variations in similarity depending on different frequencies. Therefore, a more effective feature alignment strategy is needed when aligning $\textbf{z}$ with $\textbf{z}^G$.
Thus, to align the rPPG features  across different frequencies, we introduce a regularization loss to fine-tune the extractor $F$ by,   
\begin{equation}
\label{eqn:loss_F}
\begin{split}
\mathcal{L}_F (\bm{\theta}_F) = \sum\nolimits_{
 \mathbf{z}^G \in \mathcal{B} } \lvert e^{\frac{-1}{2}(\frac{f_ \mathbf{z} - f_{\mathbf{z}^G}}{\sigma})^2} - \cos( F(\mathbf{x}; \bm{\theta}_F), \mathbf{z}^G ) \rvert,
\end{split}
\end{equation}

\noindent     
where $f_{\mathbf{z}}$ and $f_{\mathbf{z}^G}$ are frequencies in the range [0.66, 4.16] Hz corresponding to the estimated rPPG signals $E(\mathbf{z})$ and $E(\mathbf{z}^G)$, respectively; $\cos(\cdot)$ denotes the cosine function, and $\sigma$ is a predefined constant empirically set to 0.04. Figure  \ref{fig:sigma_range} illustrates the frequency-adaptive weights $e^{\frac{-1}{2} \left(\frac{f_{\mathbf{z}} - f_{\mathbf{z}^G}}{\sigma}\right)^2}$ and demonstrates how different values of $\sigma$ facilitate the weighted alignment of rPPG features across various frequencies. 


Finally, we fine-tune the model $T=E\circ F$ by, 
\begin{equation}
    \bm{\theta}_F^*,\bm{\theta}_E^* =  \mathop{\mathrm{arg}\, \mathrm{min}}_{ \bm{\theta}_F, \bm{\theta}_E} \mathcal{L}_{E}(\bm{\theta}_E) + \alpha \mathcal{L}_{F}(\bm{\theta}_F) + \beta \mathcal{L}_{T}(\bm{\theta}_F,\bm{\theta}_E), 
\label{eqn:theta_T}
\end{equation}

\noindent 
where  $\alpha$ and $\beta$ are the weight factors.
In all our experiments, we empirically set $\alpha=0.5$ and $\beta=0.1$.

\section{Experiment}



\begin{table}[]
\caption{ 
The proposed TTA-rPPG benchmark. 
}
\centering
\small
\scalebox{0.8}{ 
\begin{tabular}{c|c|c|c|c|c}
\hline
Protocols                               & Source           & Target  &Protocols                               & Source           & Target \\ \hline
\textbf{V} $\rightarrow$ \textbf{C}           &  & 45-100 bpm & \textbf{C} $\rightarrow$ \textbf{P}           &  & 40-165 bpm    \\ 
\textbf{V} $\rightarrow$ \textbf{P}           &    \multirow{2}{*}{\shortstack{45-180 \\ bpm }}                         & 40-165 bpm  &\textbf{C} $\rightarrow$ \textbf{U}           &      \multirow{2}{*}{\shortstack{45-100 \\ bpm}}                       & 55-140 bpm  \\ 
\textbf{V} $\rightarrow$ \textbf{U}           &                           & 55-140 bpm  & \textbf{C} $\rightarrow$ \textbf{V}           & & 45-180 bpm  \\ 
\textbf{V} $\rightarrow$ \textbf{[C,P,U]} &                             & 40-165 bpm & \textbf{C} $\rightarrow$ \textbf{[P,U,V]} &                             & 40-180 bpm    \\

\hline
\end{tabular}}
\label{tab:benchmark}  
\end{table}

\subsection{The proposed TTA-rPPG Benchmark 
} 
We introduce TTA-rPPG, a new fully Test-Time Adaptation benchmark for rPPG estimation, to study the challenges posed by: 1) new domain information, and 2) unbalanced HR distributions, including both seen and unseen HR ranges.  
As shown in Table~\ref{tab:benchmark}, we construct the TTA-rPPG benchmark based on four publicly available  rPPG datasets, including \textbf{COHFACE}  \cite{heusch2017reproducible} (denoted as \textbf{C}),  \textbf{PURE} \cite{stricker2014non} (denoted as \textbf{P}),   \textbf{UBFC-rPPG} \cite{bobbia2019unsupervised} (denoted as \textbf{U}), and  \textbf{VIPL-HR} \cite{niu2019rhythmnet} (denoted as \textbf{V}).
Please see the supplementary material for detailed description of the proposed TTA-rPPG benchmark.

\subsection{Evaluation Metrics and Implementation Details} 
 
We conduct experiments on the proposed TTA-rPPG benchmark and report the results using evaluation metrics, including Mean Absolute Error (MAE) and Root Mean Square Error (RMSE), across all target datasets.  
During the inference stage, we follow the method in \cite{gideon2021way,sun2022contrast} by decomposing each test video into non-overlapping 10s clips. We estimate the rPPG signals of each clip, calculate their PSDs, and then derive the corresponding HR values for each clip. 
 

 \begin{table*}[t]
\caption{
Experimental comparisons on the proposed TTA-rPPG benchmark. 
}
\centering
\footnotesize
\scalebox{1}{ 
\begin{tabular}{c|c|cccccccccccc}
\Xhline{3\arrayrulewidth}

\multirow{2}{*}{\textbf{Type}} & \multirow{2}{*}{\textbf{Method}} 
                & \multicolumn{2}{c|}{\textbf{V} $\rightarrow$ \textbf{P}}                      & \multicolumn{2}{c|}{\textbf{V} $\rightarrow$ \textbf{U}}                      & \multicolumn{2}{c|}{\textbf{V} $\rightarrow$ \textbf{C}}                      & \multicolumn{2}{c}{\textbf{V} $\rightarrow$ \textbf{[C,P,U]}} \\ 
\cline{3-10}
                & & MAE$\downarrow$ &  \multicolumn{1}{c|}{RMSE$\downarrow$} & MAE$\downarrow$ &  \multicolumn{1}{c|}{RMSE$\downarrow$} & MAE$\downarrow$ & \multicolumn{1}{c|}{RMSE$\downarrow$ } & MAE$\downarrow$ & RMSE$\downarrow$  \\
\hline
\multirow{2}{*}{DG} &
                 RErPPG-Net  (\textit{ECCV 22}) & 13.26 &  \multicolumn{1}{c|}{14.88} & 13.11 &  \multicolumn{1}{c|}{18.68} & 13.48 &\multicolumn{1}{c|}{ 15.41 } & 18.64 & 21.31 \\ 
                 & NEST (\textit{CVPR 23})  & 13.21 &  \multicolumn{1}{c|}{19.49} & 7.68 &  \multicolumn{1}{c|}{10.56} & 12.53 & \multicolumn{1}{c|}{14.60} & 16.08 & 20.38  \\ 
\hline
\multirow{1}{*}{DA} &
                 Dual-bridging (\textit{CVPR 23}) & \underline{6.58} &  \multicolumn{1}{c|}{\underline{9.16}} & \textbf{5.14} &  \multicolumn{1}{c|}{\underline{8.44}} & \underline{10.25} &  \multicolumn{1}{c|}{\underline{11.98}} & \textbf{10.25} &  \underline{12.81} \\  
\hline
\multirow{2}{*}{TTA} &
                 No adaptation & 15.26 & \multicolumn{1}{c|}{18.89 } & 14.64 &  \multicolumn{1}{c|}{18.50} &  17.19 & \multicolumn{1}{c|}{20.01} & 21.41 & 24.16 \\ 
                 & Ours  & \textbf{3.33} & \multicolumn{1}{c|}{\textbf{4.87}} & \underline{5.60} & \multicolumn{1}{c|}{\textbf{7.16}} & \textbf{8.31} & \multicolumn{1}{c|}{\textbf{9.16}} & \underline{11.14} & \textbf{12.08} \\ 

\Xhline{3\arrayrulewidth}

\multirow{2}{*}{\textbf{Type}} & \multirow{2}{*}{\textbf{Method}} 
                & \multicolumn{2}{c|}{\textbf{C} $\rightarrow$ \textbf{P}}                      & \multicolumn{2}{c|}{\textbf{C} $\rightarrow$ \textbf{U}}                      & \multicolumn{2}{c|}{\textbf{C} $\rightarrow$ \textbf{V}}                      & \multicolumn{2}{c}{\textbf{C} $\rightarrow$ \textbf{[P,U,V]}} \\ 
\cline{3-10}
                & & MAE$\downarrow$ & \multicolumn{1}{c|}{RMSE$\downarrow$} & MAE$\downarrow$ & \multicolumn{1}{c|}{RMSE$\downarrow$} & MAE$\downarrow$ & \multicolumn{1}{c|}{RMSE$\downarrow$} & MAE$\downarrow$ & RMSE$\downarrow$ \\ 
\hline
\multirow{2}{*}{DG} &
                 RErPPG-Net  (\textit{ECCV 22}) & 13.53 & \multicolumn{1}{c|}{14.16} & 16.13 &  \multicolumn{1}{c|}{17.08} & 30.52 & \multicolumn{1}{c|}{34.10} & 26.51 & 29.86 \\ 
                 & NEST  (\textit{CVPR 23}) & 16.25 & \multicolumn{1}{c|}{23.88} & 17.99 & \multicolumn{1}{c|}{22.84} & 24.90 & \multicolumn{1}{c|}{28.93} & 25.30 & 29.85 \\ 
\hline
\multirow{1}{*}{DA} & 
                 Dual-bridging (\textit{CVPR 23}) & \textbf{0.94} & \multicolumn{1}{c|}{\textbf{1.51}} & \underline{6.00} &  \multicolumn{1}{c|}{\underline{7.67}} & \underline{23.99} & 
                 \multicolumn{1}{c|}{\underline{25.86}} & \underline{25.19} & \underline{26.59}\\
\hline
\multirow{2}{*}{TTA} &
                 No adaptation & 17.93 &  \multicolumn{1}{c|}{20.12} & 19.32 & \multicolumn{1}{c|}{21.35} & 31.51 & \multicolumn{1}{c|}{34.06} & 28.72 & 30.44\\ 
                 & Ours & \underline{1.45} & \multicolumn{1}{c|}{\underline{2.24}} & \textbf{3.60} & \multicolumn{1}{c|}{\textbf{5.26}} & \textbf{12.36} & \multicolumn{1}{c|}{\textbf{13.08}} & \textbf{12.75} & \textbf{14.48} \\ 

\Xhline{3\arrayrulewidth}

\end{tabular}}
\label{tab:result}
\end{table*}

\subsection{Experimental Comparisons } 
 
Table~\ref{tab:result} presents the evaluation results of our proposed method and shows comparison with other domain generalization (DG) and domain adaptation (DA) methods on the TTA-rPPG benchmark. 
The ``No adaptation'' methods indicate that the pre-trained rPPG models $T$ remains fixed during the inference stage.
First, we observe that ``No adaptation''  yields poorer performance compared to the DG methods \cite{hsieh2022augmentation,lu2023neuron}, because a fixed pre-trained model is unable to generalize to the target domain.
Next, the DA method \cite{du2023dual}, which utilizes both labeled source data and unlabeled target data, effectively adapts the rPPG model to the target domain and achieves good performance.
However, while the DG and DA methods perform well when the HR distribution closely matches that of the source data, their performance often degrades when encountering unseen HR distributions, as shown in the protocols of \textbf{C}$\rightarrow$ \textbf{U}, \textbf{C}$\rightarrow$ \textbf{V}, and  \textbf{C}$\rightarrow$ \textbf{[P,U,V]}. In contrast, our method effectively adapts $T$ to capture HRs not present in the source training data and significantly improves the TTA performance.

\begin{table}[t]
\centering
\caption{
Ablation study on the protocols \textbf{V}$\rightarrow$ \textbf{[C,P,U]} and \textbf{C}$\rightarrow$ \textbf{[P,U,V]}, using different loss combinations.
}
\label{tab:ablation_loss}
\small
\scalebox{0.97}{ 
\centering
\begin{tabular}{ >{\centering\arraybackslash}c c c 
                | >{\centering\arraybackslash}c c
                | >{\centering\arraybackslash}c c
                }
    \hline
    \multicolumn{3}{c|}{\multirow{1}{*}{\textbf{Loss Terms}}} & 
    \multicolumn{2}{c }{\textbf{V}$\rightarrow$ \textbf{[C,P,U]}} & 
    \multicolumn{2}{c  }{\textbf{C}$\rightarrow$ \textbf{[P,U,V]}} 
    \\
    \cline{1-7}
    $\mathcal{L}_{T}$ & $\mathcal{L}_{E}$ & $\mathcal{L}_{F}$ & MAE$\downarrow$ & RMSE$\downarrow$ & MAE$\downarrow$ & RMSE$\downarrow$
    \\\hline
     -& -& -& 21.41 & 24.16 & 28.72 & 30.44 \\
    \checkmark & -&- & 17.27 & 22.29 & 21.03 & 23.54 \\
    \checkmark & \checkmark &- & 14.53 & 16.94 & 15.23 & 17.95\\
    \checkmark & \checkmark & \checkmark & 11.14 & 12.08 & 12.75 & 14.48 \\
    \hline
\end{tabular}}
\end{table}

\begin{figure}[t]
    \begin{minipage}[t]{1\columnwidth}
      \centering{ 
      \includegraphics[width=0.8\columnwidth]{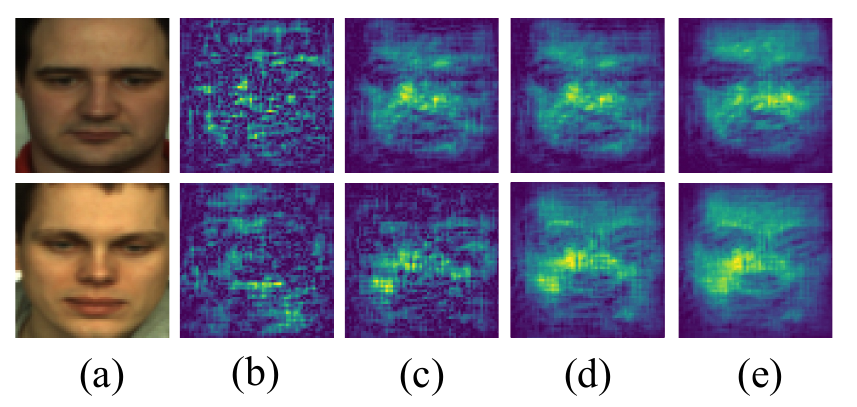}
      }
    \end{minipage}   
\caption{    
Two examples of (a) input facial images and (b)-(e) the saliency maps \cite{simonyan2013deep} generated by $T$ trained with different loss terms listed in Table~\ref{tab:ablation_loss}. For an effective rPPG estimator, the visualized saliency maps should show a strong response in the facial skin regions.
}
 \label{fig:saliency}   
\end{figure}

\subsection{Ablation Study}

\subsubsection{Comparison between Different Loss Terms}
 
In Table~\ref{tab:ablation_loss}, we compare using different combinations of loss terms to update $T= E \circ F$ during the online training (adaptation) process of inference stage. 
We test on the protocols \textbf{V}$\rightarrow$ \textbf{[C,P,U]} and \textbf{C}$\rightarrow$ \textbf{[P,U,V]}, as shown in Table \ref{tab:benchmark}, and 
visualize the saliency maps of two examples in Figure~\ref{fig:saliency}.

 
First, we use the fixed pre-trained model without any online training (adaptation) process as the baseline.  
We observe that the baseline model performs poorly, and the saliency maps in Figure~\ref{fig:saliency}(b) appear unclear, likely due to the presence of new domain information and unbalanced HR distributions with unseen HR ranges in the target data. 
 
Next, we include $\mathcal{L}_{T}$ to fine-tune $F$ and $E$.
We see that the proposed spectral-based entropy minimization effectively adapts the model $T$ to learn the target domain information.
The saliency maps in Figure~\ref{fig:saliency} (c) also show improvement compared to Figure~\ref{fig:saliency} (b). 
  
Moreover, we include $\mathcal{L}_{E}$ and use the paired $  (\widetilde{\mathbf{s}},\mathbf{z}^G)$ stored from  $\mathcal{B}$ to enhance the adaptation of $E$.
Because the generator $G$ guided by the synthetic rPPG signals $\widetilde{\mathbf{s}}$ is able to generate latent rPPG features capable of producing the estimated rPPG signals similar to $\widetilde{\mathbf{s}}$, we see that the generated $  (\widetilde{\mathbf{s}},\mathbf{z}^G)$ indeed facilitates the adaptation of $E$  to prevent the forgetting issue caused by unbalanced HR distributions, but also to effectively  measure the HRs unseen within the source data.  
In addition, we see the saliency maps in Figure~\ref{fig:saliency} (d) more accurately concentrate on facial regions compared to Figure~\ref{fig:saliency} (c).   
 
Finally, when including $\mathcal{L}_{F}$, we achieve the best performance and saliency maps in Figure~\ref{fig:saliency} (e) by continually aligning the extracted rPPG features $\mathbf{z}^G$ to the generated rPPG features $\mathbf{z}^G$.
The overall improvement verifies the superiority of the proposed synthetic signal-guided feature learning.

\begin{table}[t]
\centering
\caption{ 
Ablation study on the protocols \textbf{V}$\rightarrow$ \textbf{[C,P,U]} and \textbf{C}$\rightarrow$ \textbf{[P,U,V]},
using different optimization strategies. 
}
\label{tab:ablation_update}
\setlength{\tabcolsep}{3pt}
\small
\scalebox{0.95}{ 
\centering
\begin{tabular}{ >{\centering\arraybackslash}c
                | >{\centering\arraybackslash}c c
                | >{\centering\arraybackslash}c c
                | c}
    \hline
    \multicolumn{1}{c|}{\multirow{2}{*}{\textbf{\shortstack{Updating \\ parameters}}}} & 
    \multicolumn{2}{c|}{\textbf{V}$\rightarrow$ \textbf{[C,P,U]}} & 
    \multicolumn{2}{c|}{\textbf{C}$\rightarrow$ \textbf{[P,U,V]}} & \multirow{2}{*}{\textbf{FPS}} \\
    \cline{2-5}
    & MAE$\downarrow$ & RMSE$\downarrow$ & MAE$\downarrow$ & RMSE$\downarrow$
    \\\hline
    BatchNorm layers & 15.85 & 19.75 & 20.14 & 21.71 & 1146.15\\
    Entire model $T$ & 11.14 & 12.08 & 12.75 & 14.48 &912.12 \\
    \hline
\end{tabular}}
\end{table}

\begin{table*}[t]
\centering
\caption{ 
\textcolor{black}{
HRV and HR estimation results on UBFC-rPPG. 
}
}
\label{tab:HRV}
\setlength{\tabcolsep}{3pt} 
\renewcommand{\arraystretch}{1.1}
\scalebox{0.85}{ 
\centering
\begin{tabular}{ >{\centering\arraybackslash}c 
                | >{\centering\arraybackslash}c c c
                | >{\centering\arraybackslash}c c c
                | >{\centering\arraybackslash}c c c
                | >{\centering\arraybackslash}c c c
                | >{\centering\arraybackslash}c c}
    \hline
    \multirow{2}{*}{\textbf{Method}} & 
    \multicolumn{3}{c|}{LF (n.u.)} & \multicolumn{3}{c|}{HF (n.u.)} & \multicolumn{3}{c|}{LF/HF} & \multicolumn{3}{c|}{RF(Hz)} & \multicolumn{2}{c}{HR(bpm)}  \\
    \cline{2-15}
     & 
    STD$\downarrow$ & RMSE$\downarrow$ & R$\uparrow$ & STD$\downarrow$ & RMSE$\downarrow$ & R$\uparrow$ & STD$\downarrow$ & RMSE$\downarrow$ & R$\uparrow$ & STD$\downarrow$ & RMSE$\downarrow$ & R$\uparrow$ & MAE$\downarrow$ & RMSE$\downarrow$ 
    \\\hline   
    RErPPGNet (\textit{ECCV 22}) & 0.128 & 0.194 & 0.667 & 0.128 & 0.194 & 0.667 & 0.452 & 0.603 & 0.496 & 0.050 & 0.060 & 0.011 & 2.00 & 2.40 \\
    NEST (\textit{CVPR 23}) & 0.131 & 0.216 & 0.259 & 0.131 & 0.216 & 0.259 & 0.539 & 0.800 & 0.236 & 0.048 & 0.060 & 0.158 & 1.29 & 4.97 \\
    Dual-bridging (\textit{CVPR 23}) & 0.138 & 0.172 & 0.397 & 0.138 & 0.172 & 0.397 & 0.500 & 0.556 & 0.227 & 0.069 & 0.087 & 0.187 &
    \textbf{0.16 } &  \textbf{0.57 }  \\
    \hline
    No adaptation & 0.161 & 0.227 & 0.291 & 0.161 & 0.227 & 0.291 & 0.592 & 0.862 & 0.242 & 0.116 & 0.152 & 0.174 & 2.84 & 3.68 \\
    Ours  & \textbf{0.055}  & \textbf{0.103}   & \textbf{0.771 }  & \textbf{0.055}   & \textbf{0.103}   & \textbf{0.771}   & \textbf{0.167}   &\textbf{0.219}   & \textbf{0.790}  &\textbf{0.062}   & \textbf{0.123}  & \textbf{0.270}  & 0.40  & 0.69  \\
    \hline
\end{tabular}}
\end{table*}

\subsubsection{Comparison of Different Optimization Strategies} 
 
In Table~\ref{tab:ablation_update}, we compare using different optimization strategies to fine-tune $T$ under the same loss $\mathcal{L}_{T} + \mathcal{L}_{E}+ \mathcal{L}_{F}$
during the online training (adaptation) process.
We first refer to previous classification-based TTA methods \cite{wang2021tent, park2023label} that only update the batch normalization (BN) layers.
We see that updating BN layers enables rapid response and aids $T$ in learning target domain information.
However, updating only BN layers is insufficient to effectively adapt $T$ to learn unseen HR ranges, resulting in unsatisfactory performance.  
By contrast, when updating the entire model $T$, the significant improvement demonstrates that $T$ better adapts to the target data  
new domain information, unbalanced HR distributions, and  unseen HR ranges without encountering significant speed degradation.


\begin{table}[t]
\centering
\caption{ Ablation study on the protocols  \textbf{V}$\rightarrow$ \textbf{[C,P,U]} and \textbf{C}$\rightarrow$ \textbf{[P,U,V]}, using different feature alignments (FA).}
\label{tab:ablation_MB_weighting}
\small
\scalebox{1}{ 
\centering
\begin{tabular}{ >{\centering\arraybackslash}c
                | >{\centering\arraybackslash}c  c 
                | >{\centering\arraybackslash}c  c }
    \hline
    \multirow{2}{*}{\textbf{Methods}} & 
    \multicolumn{2}{c}{\textbf{V}$\rightarrow$ \textbf{[C,P,U]}} & 
    \multicolumn{2}{c}{\textbf{C}$\rightarrow$ \textbf{[P,U,V]}}  \\
    \cline{2-5}
     &
    MAE$\downarrow$ & RMSE$\downarrow$ & 
    MAE$\downarrow$ & RMSE$\downarrow$ 
    \\\hline
    Dichotomous FA  & 13.85 & 15.88& 14.71 & 16.25\\
    Adaptive FA & 11.14 & 12.08 & 12.75 & 14.48 \\
    \hline
    
\end{tabular}}
\end{table}

\begin{table}[t]
\centering
\caption{ 
Ablation study on the protocols   \textbf{V}$\rightarrow$ \textbf{[C,P,U]} and \textbf{C}$\rightarrow$ \textbf{[P,U,V]}, using different memory banks (MB).
}
\label{tab:ablation_MB_design}
\small
\scalebox{0.95}{ 
\centering
\begin{tabular}{ >{\centering\arraybackslash}c
                | >{\centering\arraybackslash}c  c 
                | >{\centering\arraybackslash}c  c }
    \hline
    \multirow{2}{*}{\textbf{Methods}} & 
    \multicolumn{2}{c}{\textbf{V}$\rightarrow$ \textbf{[C,P,U]}} & 
    \multicolumn{2}{c}{\textbf{C}$\rightarrow$ \textbf{[P,U,V]}}  \\
    \cline{2-5}
     &
    MAE$\downarrow$ & RMSE$\downarrow$  & 
    MAE$\downarrow$ & RMSE$\downarrow$  
    \\\hline
    Vanilla MB & 13.18 & 14.42 & 14.78 & 15.50 \\
    Frequency-based MB & 11.14 & 12.08 & 12.75 & 14.48 \\
    \hline
\end{tabular}}
\end{table}

\begin{table}[t]
\caption{ 
\textcolor{black}{
Ablation study on different network architectures.
}
}
\label{tab:transferability}
\setlength{\tabcolsep}{2.5pt} 
\centering
\small
\scalebox{0.77}{ 
\centering
\begin{tabular}{ >{\centering\arraybackslash}c 
                | >{\centering\arraybackslash}c
                | >{\centering\arraybackslash}c c c c c }
    \hline
    \multirow{2}{*}{\textbf{Network $T$}} & \multirow{2}{*}{\textbf{Method}} & 
    \multicolumn{5}{c}{\textbf{C}$\rightarrow$ \textbf{[P,U,V]}} \\
    \cline{3-7}
     & & 
    MAE$\downarrow$ & RMSE$\downarrow$  & Model size & GFLOPs & FPS 
    \\\hline
    \multirow{2}{*}{PhysNet  } & No adaptation & 28.72 & 30.44 & \multirow{2}{*}{2.93MB} & \multirow{2}{*}{196.72} & 4327.01 \\
     & Ours  & 12.75 & 14.48 &  &  & 912.12  \\
    \hline
    \multirow{2}{*}{PhysFormer  } &  No adaptation & 24.16 & 28.67 & \multirow{2}{*}{28.15MB} & \multirow{2}{*}{296.06} & 1597.051 \\
     & Ours & \textbf{11.48} & \textbf{12.71} &  &  & 464.87  \\
    \hline
\end{tabular}}
\end{table}

\subsubsection{Heart Rate Variability Analysis}
In Table~\ref{tab:HRV}, we conduct heart rate variability (HRV) and heart rate (HR) estimation on intra dataset (i.e., \textbf{U}) to explore whether the improvement in heart rate measurement efficacy comes at the expense of reduced rPPG waveform quality.
The results in Table~\ref{tab:HRV} and Table 6 all demonstrate the effectiveness of the proposed method in both intra-domain and cross-domain testing.
In addition, our method also produces better quality rPPG waveforms to improve the HR measurement.

\subsubsection{Comparison of Different Feature Alignment Strategies}  
In Table~\ref{tab:ablation_MB_weighting}, we compare using different feature alignments (FA) to align the extracted rPPG features $\textbf{z}=F(\mathbf{x})$ and the generated rPPG features  $\textbf{z}^G$.
First, we adopt the dichotomous feature alignment \cite{gideon2021way,sun2022contrast} to align $\mathbf{z}=F(\mathbf{x})$ and $\mathbf{z}^G$ from  \textbf{the same HR} and to push $\mathbf{z}=F(\mathbf{x})$ and $\mathbf{z}^G$ from \textbf{different HRs}.
Next, by adopting the proposed adaptive feature alignment to align $\mathbf{z}=F(\mathbf{x})$ and $\mathbf{z}^G$ from the \textbf{similar HRs}, the adaptation of $T$ is improved.
This performance enhancement also coincides with the finding observed in \cite{yang2023simper} that learning similar features of periodic signals with similar frequencies (i.e., HRs) indeed facilitates rPPG estimation.

 \begin{figure}[t] 
    \centering 
     \includegraphics[width=1\columnwidth]{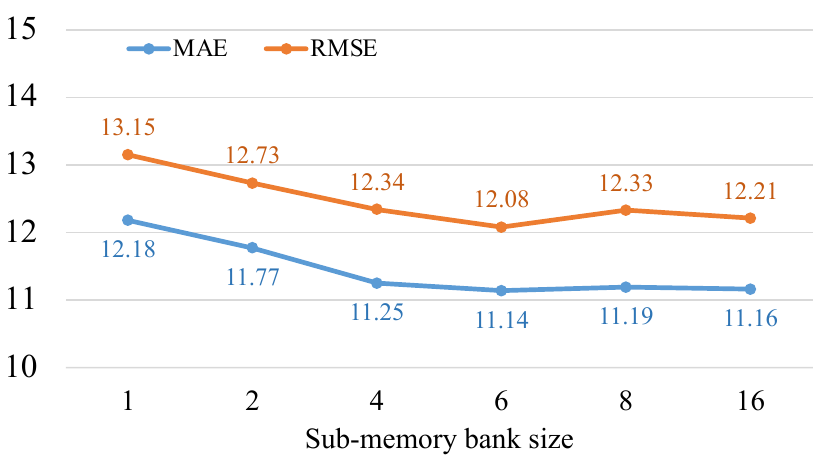} 
\caption{ 
Ablation study on the protocol \textbf{V}$\rightarrow$ \textbf{[C,P,U]}, using different sub-memory bank sizes. 
}
 \label{fig:MB_size}  
  \vspace{-0.35cm}
\end{figure}

\subsubsection{Comparison of  Different Memory Banks}
 
In Table~\ref{tab:ablation_MB_design}, we compare using different memory banks (MB) of the same total size to store the generated $  (\widetilde{\mathbf{s}},\mathbf{z}^G)$ under the same First In First Out (FIFO) scheme.
Because the vanilla MB adopts the FIFO scheme during the training of $G$, it cannot guarantee that the stored pairs $(\widetilde{\mathbf{s}},\mathbf{z}^G)$ cover the entire HR range.
By contrast, the proposed frequency-based MB  ensures that the stored pairs $(\widetilde{\mathbf{s}},\mathbf{z}^G)$ cover the entire HR range to effectively tackle the forgetting issue.

\subsubsection{Comparison of Different Memory Bank Sizes}
 
In Figure~\ref{fig:MB_size}, we compare using different sub-memory bank sizes $N_{\mathcal{B}}$ ($N_{\mathcal{B}}$ = 1, 2, 4, 6, 8 and 16)  in $\mathcal{B}$ on the protocol \textbf{V}$\rightarrow$ \textbf{[C,P,U]} and show the results in terms of MAE and RMSE.
First, We observe that the performance steadily improves  as $N_{\mathcal{B}}$ increases from 1 to 6 and reaches the best performance at  $N_{\mathcal{B}}=6$. 
These results demonstrate that larger sub-memory banks indeed facilitate better adaptation of the rPPG models to unseen HR distributions.
Next, we see that the performance does not improve as  $N_{\mathcal{B}}$ increases from 8 to 16.
This performance plateau might indicate that the memory bank has reached a state of overcompleteness.
Therefore, based on this ablation study, we empirically set $N_{\mathcal{B}}=6$ for subsequent experiments.

\begin{figure}[t] 
    \centering 
     \includegraphics[width=1\columnwidth]{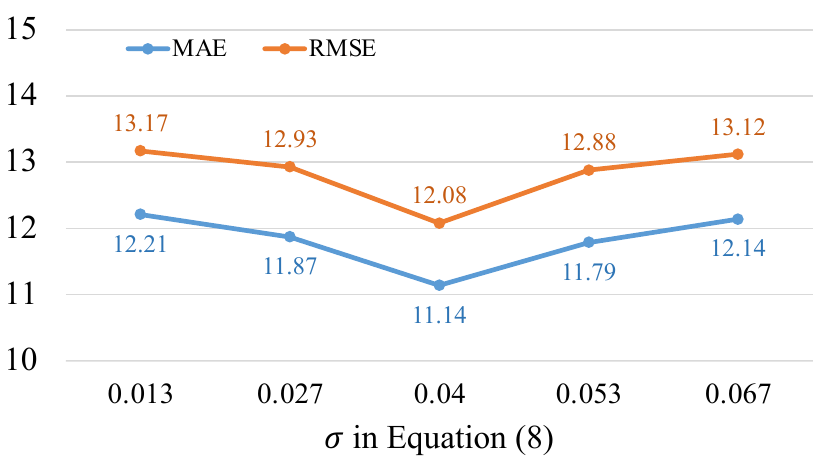} 
\caption{ 
Ablation study on the protocol \textbf{V}$\rightarrow$ \textbf{[C,P,U]}, using different $\sigma$ in Equation \eqref{eqn:loss_F}.
}
 \label{fig:sigma}   
\end{figure}

\subsubsection{Comparison of Different $\sigma$ Values}
 
In Figure~\ref{fig:sigma}, we compare using different $\sigma$ in Equation \eqref{eqn:loss_F} to align similar rPPG features from different HR ranges. 
Note that, different values of $\sigma$ corresponding to valid HR ranges are given in Figure \ref{fig:sigma_range}. 
Because rPPG features corresponding to similar HRs exhibit proximity in the latent feature space \cite{yang2023simper}, 
we see that aligning rPPG features within HR range of minor difference (i.e., $\sigma \leq 0.04$, corresponding to HR difference of less than 5) does indeed facilitate the feature alignment process, as shown in Figure~\ref{fig:sigma}. 
In addition, when aligning rPPG features within HR range of larger difference, their corresponding rPPG features exhibit differences that hinder effective feature alignment.

\subsubsection{Comparison of Different Network Architectures}
In Table~\ref{tab:transferability}, we investigate the impact of different network architectures (including PhysNet \cite{yu2019physnet} and PhysFormer \cite{yu2022physformer}) using the same loss $\mathcal{L}_{T} + \mathcal{L}_{E}+ \mathcal{L}_{F}$ on the most challenging protocol \textbf{C} $\rightarrow$ \textbf{[P,U,V]}.
The results in Table~\ref{tab:transferability} verify robust transferability of the proposed method using different network architectures. 
Moreover, the results in Table~\ref{tab:transferability} show that, larger and more powerful network architectures indeed yield better performance but have lower inference speed due to the increased number of parameters.


\section{Conclusion}

In this paper, we introduce a new benchmark TTA-rPPG and propose a novel synthetic signal-guided generative feature learning approach to address challenges of TTA in rPPG estimation. 
First, we propose a novel synthetic signal-guided generative feature learning to generate and store supplementary rPPG information  to address the issues of forgetting and unbalanced HR distributions. 
Next, we also design a spectral-based entropy minimization mechanism to adapt the rPPG models to new domain. 
Finally, we propose an effective feature-aligned rPPG estimation to align the feature distributions between the target rPPG features and the generated rPPG feature to further enhance the model’s ability to adapt to the characteristics of the target data during inference. 
Our experimental results on the proposed benchmark TTA-rPPG demonstrate the superior performance of the proposed method on handling realistic TTA scenarios.

\bibliographystyle{IEEEtran}
\bibliography{egbib} 

\end{document}